\documentclass{article}

\usepackage[final]{corl_2019} %
\usepackage{float}
\usepackage{algorithm}

\usepackage[utf8]{inputenc} %
\usepackage[T1]{fontenc}    %
\usepackage{hyperref}       %
\usepackage{url}            %
\usepackage{booktabs}       %
\usepackage{amsfonts}       %
\usepackage{nicefrac}       %
\usepackage{microtype}      %

\usepackage{tikz}
\usepackage{algorithm}
\usepackage{algpseudocode}
\usepackage{multicol}
\usepackage{enumitem}
\setlist{nolistsep,noitemsep,itemindent=-24pt}
\usepackage{wrapfig}
\usepackage{subcaption}

\usepackage[size=small]{caption}
\usetikzlibrary{automata, positioning}
\usepackage{amsmath, amssymb}

\newcommand{\vv}{\mathbf{v}}    %
\newcommand{\Em}{\mathbf{E}}
\newcommand{\Ecnn}{\mathbf{E}_\text{CNN}}

\newcommand{\Ecnnon}{\mathbf{E}_\text{CNN}(o_{t+1})}
\newcommand{\Egnn}{\mathbf{E}_\text{GNN}}
\newcommand{\Egnns}{\mathbf{E}_\text{GNN}(s_t)}
\newcommand{\Egnnsn}{\mathbf{E}_\text{GNN}(s_{t+1})}
\newcommand{\Egnng}{\mathbf{E}_\text{GNN}(g)}
\newcommand{\Loss}{\mathcal{L}}

\newcommand{\phio}{\phi_{o_t}}
\newcommand{\phion}{\phi_{o_{t+1}}}
\newcommand{\phig}{\phi_{g}}
\newcommand{\phigo}{\phi_{g^o}}
\newcommand{\phis}{\phi_{s_t}}
\newcommand{\phisn}{\phi_{s_{t+1}}}
\newcommand{\xx}{\mathbf{x}}
\newcommand{\pp}{\mathbf{p}}

\algnewcommand{\LineComment}[1]{\(\triangleright\) #1}
\def\globals{\mathbf{u}}
\newcommand{\Lstate}{\mathcal{L}_\text{STATE}}
\newcommand{\Ldyn}{\mathcal{L}_\text{DYN}}
\newcommand{\Lfull}{\mathcal{L}_\text{FULL}}
\newcommand{\footurl}[1]{\footnote{\url{#1}}}

\usepackage{fixme}
\fxsetup{status=draft, theme=color}
\definecolor{fxtarget}{rgb}{0.8000,0.0000,0.0000}
\definecolor{fxnote}{rgb}{0.0000,0.0000,0.8000}

\title{Learning to Manipulate Object Collections\\Using Grounded State Representations}

\author{
  Matthew Wilson\\
  University of Utah\\
  United States\\
  \texttt{matthew.b.wilson@utah.edu}
  \And
  Tucker Hermans \\
  University of Utah \& NVIDIA\\
  United States\\
  \texttt{thermans@cs.utah.edu}
}

\begin{document}
\maketitle

\begin{abstract}
	We propose a method for sim-to-real robot learning which exploits simulator 
	state information in a way that scales to many objects.
	We first train a pair of encoder networks to capture multi-object state information in a latent space.
	One of these encoders is a CNN, which enables our system to operate on RGB images in the real world; the other is a graph neural network (GNN) state encoder, which directly consumes a set of raw object poses and enables more accurate reward calculation and value estimation.
	Once trained, we use these encoders in a reinforcement learning algorithm to train image-based policies that can manipulate many objects.
	We evaluate our method on the task of pushing a collection of objects to desired tabletop regions.
	Compared to methods which rely only on images or use fixed-length state 
	encodings, our method achieves higher success rates, 
	performs well in the real world without fine tuning, and generalizes
	to different numbers and types of objects not seen during training.
	Video results: \href{https://bit.ly/2khSKUs}{bit.ly/2khSKUs}.
\end{abstract}


\section{Introduction}
\label{sec:intro}

Humans regularly manipulate object groups as collections.
When scooping up a bunch of grapes or sweeping a pile of coins into their hands,
humans can track and manipulate the objects without needing to know, for example, how many nickels or dimes are present.
Current robot systems lack such capabilities and most recent work has focused on picking or pushing objects one at a time~\citep{tella-trcyb1982-bin-picking,mason1985,berenson-ichr2008,cosgun-iros2011,hermans-iros2012,mahler-scirob2019-dexnet4}.
Learning to simultaneously manipulate many objects, on the other hand, is a greater challenge, as it requires a robot to track and reason about the many possible 
configurations and physical interactions between objects.
To overcome this challenge, we use raw object poses from a simulator 
to learn a latent space that captures multi-object state information; we then use this learned representation to
train an image-based policy that reasons about and manipulates variably-sized
object collections as a whole.

Learning policies directly from RGB images stands as a popular approach for solving
manipulation tasks, but as an approach it comes with several challenges.
Namely, it is not obvious how to form task-relevant features
or generate meaningful reward functions directly from high-dimensional sensory inputs.
Since labeled data for manipulation requires significant time and effort to collect,
researchers often leverage indirect self-supervised signals for training (e.g., \citep{finn16, rig, pokebot}).
Since such signals, by definition, do not directly correspond to task-relevant state information
(such as poses and velocities of relevant objects), they can be prone to failure.
For example, an autoencoder loss \citep{finn16, rig} will incentivize a network
to waste latent space capacity modeling more visually salient large objects and background in a scene, sometimes ignoring more fine-grained information relevant to the task.%

\begin{figure}[t]
	\centering
	\includegraphics[width=\linewidth]{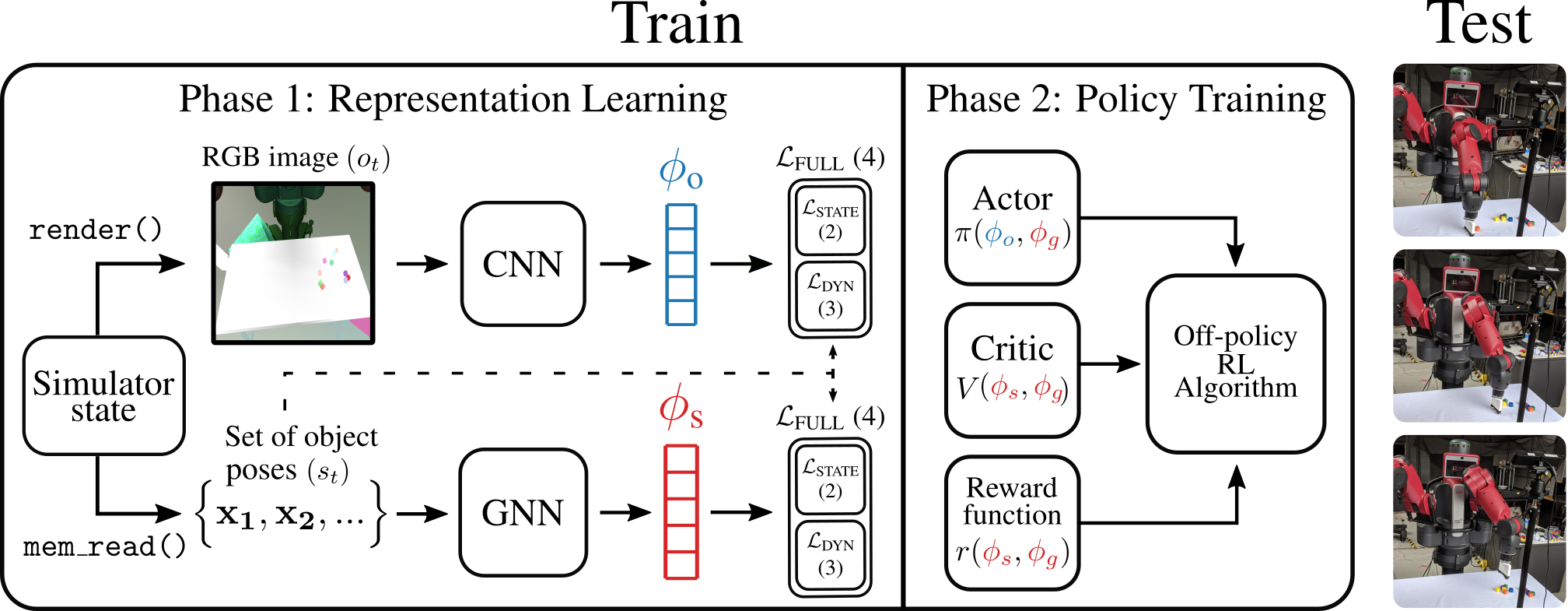}
	\caption[cartoon]{{Cartoon diagram of our approach. 
	We first independently train two encoder networks, one convolutional neural network (CNN) and one graph neural network (GNN) using
	a multi-object state and dynamics loss function.
	Then, during our RL phase, we embed the observation: $o \xrightarrow{\text{CNN}} \phi_o$, 
	state: $s \xrightarrow{\text{GNN}} \phi_s$, and goal: $g \xrightarrow{\text{GNN}} \phi_g$,
	and we use the embeddings in an asymmetric actor critic framework \citep{aac}
	to train a multi-object policy $\pi$.%
	}}
	\label{fig:overview}
\vspace{-0.18in}
\end{figure}

Simulation-based training is a promising route to sidestep these difficulties.
Researchers have shown that by exploiting raw state information
to calculate rewards \citep{aac, openai18, peng18, xie19, anymal, minotaur, matas18, rcan, chebotar-icra2019}, construct grounded latent representations \citep{aac, openai18, zhang16}, and improve value function
estimation \citep{aac, openai18}, they can greatly improve the stability and speed of policy training.  %
However, all existing work rely on \emph{fixed-length} vector inputs,
limiting their effective use to settings with a known number of individual objects. 
Ours is the first work we are aware of in this space capable of handling a variable number of multiple objects.

Our primary contribution is an approach for exploiting simulator state information in a way that scales to multiple and variable object settings.
In our approach, we train a pair of encoder networks to capture
the state and dynamics information of a variable number of objects;
we then use them to aid in learning multi-object manipulation policies.
Our pair of encoders consists of one convolutional neural network (CNN)
trained with RGB image inputs on sets of object pose targets
and one graph neural network (GNN)~\citep{graph_nets} trained with 
sets of object poses as both its inputs and targets (i.e., it learns to autoencode the state).
Since the CNN operates directly on RGB images, it gives us the ability to learn a policy that can be deployed in the real world.
Since the GNN is doubly grounded in simulator state information (via input and output),
it provides a more accurate and stable representation, so we use
it both for calculating rewards and feeding to the value network
in our actor-critic RL algorithm.
(The idea of providing different inputs to the policy
and value networks in this way is known as an asymmetric actor critic \citep{aac}.)
Fig.~\ref{fig:overview} provides an overview of our network structure and full approach.

To evaluate our approach, we construct a multi-object manipulation task ({Fig.~\ref{fig:task}) which is challenging
to learn from RGB images alone and ill-suited for prior fixed-state representation learning approaches.
We consider a collection of 1-20 homogeneously shaped objects that start in arbitrary
configurations on a tabletop, which the robot must push into desired goal configurations.
We train on a simulated version of this task but we learn a policy that can be deployed in the real-world without any fine tuning.
We conduct extensive comparison and ablation studies of our trained policy in both simulation and on the physical robot.
We find that our approach achieves higher success rates than an autoencoder or a fixed-state
training approach, while also generalizing well to configurations not seen during training.

\section{Related Work}
\label{sec:relwork}
\vspace{-0.75em}
We draw on and extend two main lines of robot learning research: self-supervised representation learning and sim-to-real learning.

\paragraph{Self-Supervised Representation Learning:}
A common pattern in robot learning methods
is to pre-train a representation using a self-supervised signal and
then use that representation for downstream task planning or task learning 
\citep{finn16, rig, pokebot, rcan, ropebot, ebert17, ebert18, pinto15, solar, Lee2019c, sutanto-icra2019, graspgan}.
The pattern usually goes as follows:
researchers use an exploration policy (often scripted \citep{rig, pokebot, ropebot, ebert17, ebert18})
to collect a dataset of real world interactions, consisting of either
single frames \citep{finn16, rig},
state action transitions \citep{pokebot, ropebot, solar},
or long sequences of frames and actions \citep{ebert17, ebert18}.
They then use this dataset to train a model via a self-supervised signal, such 
as an autoencoder loss \citep{finn16, rig}, GAN loss \citep{rcan, graspgan}, 
or some related pre-text task~\citep{pokebot, ropebot, solar, Lee2019c, se3, se3pose}.
Once they learn a good representation, they often use it in either
a model-based/planning framework \citep{pokebot, ropebot, ebert17, ebert18, solar, sutanto-icra2019,se3pose}
or in a model-free RL framework \citep{finn16, rig} to learn a task policy.
In this work, we follow a model-free RL approach.
Compared to prior work, we learn representations which are \emph{grounded},
both in their inputs and outputs.
Unlike most prior work, where representations are trained on image-based predictions, 
we directly train our representations to capture the state of the objects (through Eq.~\ref{eq:mdn-loss}).
We also directly consume ground truth information
via a GNN~\citep{graph_nets}.
Our experiments suggest that these components speed
up and stabilize policy training and improve the quality
and generalizability of trained models.

\paragraph{Sim-to-Real Learning:}
Sim-to-real methods---those trained primarily or exclusively in physics
simulators and targeted to work in the real world---are starting to yield
impressive results on physical robotics tasks \citep{aac, openai18, peng18, xie19, anymal, minotaur, matas18, rcan, chebotar-icra2019, graspgan, sadeghi16, tobin17}.
Simulation training offers several advantages
over real-world training, including increased
speed and parallelizability of data collection,
safety of exploration, and full observability and
control over environmental factors.
However, there is often a large gap between the crude
simulator phenomena and their real world analogs
which causes naively-trained models to fail in the real world.
Researchers have developed several methods to help enable
policies to transfer across this gap,
including using GANs to map from simulation visuals 
to the real world~\citep{graspgan} or vice versa~\citep{rcan},
and incorporating learned models from the real world
into the simulation loop \citep{anymal}.
One simple and effective approach we use, known as domain randomization~\cite{tobin17}, is to randomize
the physical and visual properties of an environment
during training.
This variability makes trained policies more robust
and less prone to overfitting to simulator characteristics \citep{aac, peng18, minotaur, sadeghi16, tobin17}.
In this work, we demonstrate a sim-to-real approach for speeding up training in multi-object settings.
Our approach is symbiotic with existing methods, scales to a variable number of objects,
and is extensible to more complex tasks through its flexible state encoding.

\paragraph{Graph Neural Networks in RL:}
GNNs are becoming popular in control and decision making after their strong performance
in natural language processing \cite{transformer} and visual-spatiability tasks \cite{rns}.
DeepMind researchers use GNNs in RL to 
play the Star Craft II video game \citep{zambaldi18, alphastar}.
\citet{ajay19} use a GNN for robotic manipulation 
to create a hybrid analytic and data-driven simulator for task planning; 
however, they rely on precise object knowledge and motion capture in the real world.
Our work is the first we know of to use GNNs to learn state
representations for robotic manipulation tasks.

\section{Preliminaries}
\label{sec:prelims}

\paragraph{Goal-conditioned RL:}
We use the standard multi-goal or goal-conditioned reinforcement learning formalism, as in \citet{uvfa}.
We describe an environment by states $s \in \mathcal{S}$, goals $g \in \mathcal{G}$,
actions $a \in \mathcal{A}$,
a distribution for sampling initial state-goal pairs $p(s_0, g)$, and a goal-conditioned
reward function $r: \mathcal{S} \times \mathcal{A} \times \mathcal{G} \rightarrow \mathbb{R}$.
To account for partial observability, we denote images as observations $o \in \mathcal{O}$,
and images of goals as $g^o \in \mathcal{O}$.  The objective of the agent is to learn a
goal-conditioned policy $a_t = \pi(o_t, g)$ which maximizes the expected discounted return $\mathbb{E}[R_t] = \mathbb{E}[\sum_{i=t}^\infty \gamma^{i-t} r_i]$.

\begin{wrapfigure}{r}{.4\columnwidth}
\vspace{-8mm}
\centering
\includegraphics[width=0.37\columnwidth]{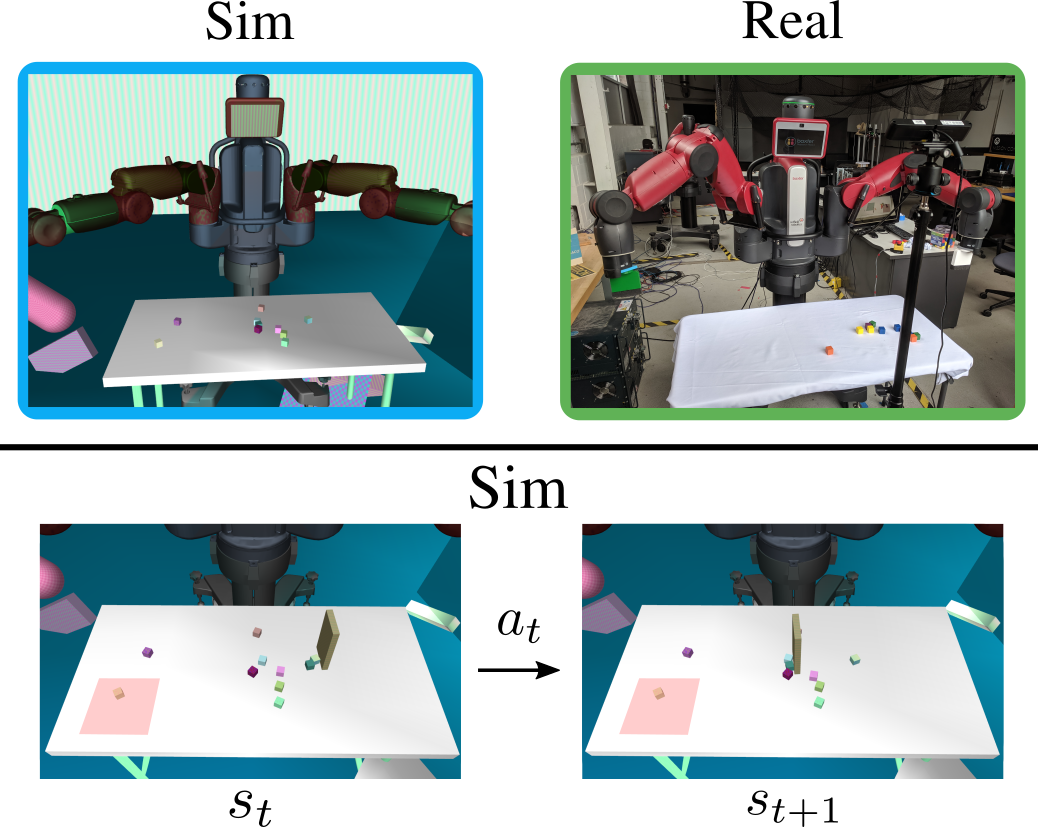}
\vspace{-0.5mm}
	\caption{\textbf{Top:} Simulation and real world environments.
	\textbf{Bottom:} An example of a state-action transition in simulation
	showing a trained policy pushing towards a goal area,
	which is marked by a red square.
	}
\label{fig:task}
\vspace{-8mm}
\end{wrapfigure}

\paragraph{Task Overview:}
The task we evaluate on
requires an agent to push a collection of 1-20 homogeneous objects
to desired areas on a tabletop.
The agent's state is factored into a set of 2D coordinates of object centers
$\{\xx_1, \xx_2, ...\}$.
Similar to prior work \cite{pokebot},
we use pushing actions which are parameterized
by 4 continuous values: a 2D starting pose  $(x,y,\theta)$ and a pushing distance \(d\).
These values are constrained to the reachable table workspace of
our physical Baxter robot \cite{baxter}.
For data collection we use a simple scripted policy, which
randomly samples an object to push for a random distance in a random direction.

\paragraph{Graph Neural Networks:}
We use a simplified version of the graph neural network framework introduced by \citet{graph_nets}.
We define graphs by a 3-tuple $G = (V, E, \globals)$,
where $V = \{\vv_i\}$ is the set of nodes with each $\vv_i$ a
vector representing the 2D pose of an object; $E = \{(\vv_r, \vv_s)\}$
is an edge between a pair of receiver, $\vv_r$, and sender, $\vv_s$, nodes;
and $\globals$ is a global attribute vector which
we compute to aggregrate the full graph information.
GNNs have been used in a variety of domains \cite{rns, ajay19} and
they vary widely in how they operate on graph structures \cite{graph_nets}.
However, they all use some variation of two key functions: \emph{updates} and \emph{aggregations} \cite{graph_nets}.
Updates run computations on individual nodes or edges (e.g., linear layer forward pass on each node).
Aggregrations perform reductions across graph structures
(e.g., elementwise summation over the set of all nodes).
See \citet{graph_nets} for a more in-depth introduction to GNNs.

\section{Learning to Manipulate Object Collections}
\label{sec:method}
\vspace{-0.75em}
To train an image-based policy capable of tracking and 
reasoning about the many possible configurations and interactions in 
multi-object tasks, we develop a method to exploit variably-sized state information 
from a physics simulator.

Our approach, illustrated in Fig.~\ref{fig:overview}, consists of two phases:
a supervised representation learning phase and a reinforcement learning (RL) 
policy training phase.
The interesting details of our apporach lay mostly in how we train our
representations to capture variably-sized information
and how we incorporate them in the RL phase to aid in policy training.
Subsection~\ref{sec:gsr} covers the supervised learning phase and
our solutions to the technical challenges of consuming and predicting
variably-sized object state information (Fig.~\ref{fig:arch}).
Subsection~\ref{sec:rl} covers the RL phase and how we calculate rewards
and incorporate our learned representations into a full policy-training algorithm.

\vspace{-0.5em}
\subsection{Learning Grounded State Representations}
\vspace{-0.5em}
\label{sec:gsr}
We seek to learn representations which capture the variably-sized and order-invariant \emph{set} of objects in the environment state \(s\).  Formally, we seek to learn two encoders \(\Em_s(s)\) and \(\Em_o(o)\) that map from their respective modalities to latent vector spaces \(\phi_s\) and \(\phi_o\) which each represent the relevant information of the variably-sized state.  This objective introduces challenges both on encoding the inputs and decoding the outputs of the neural networks. We discuss these challenges and our solutions below.

\vspace{-0.25em}
\paragraph{The Input Side - Encoding Raw State:}
\begin{figure}[t]
	\centering
	\includegraphics[width=\linewidth]{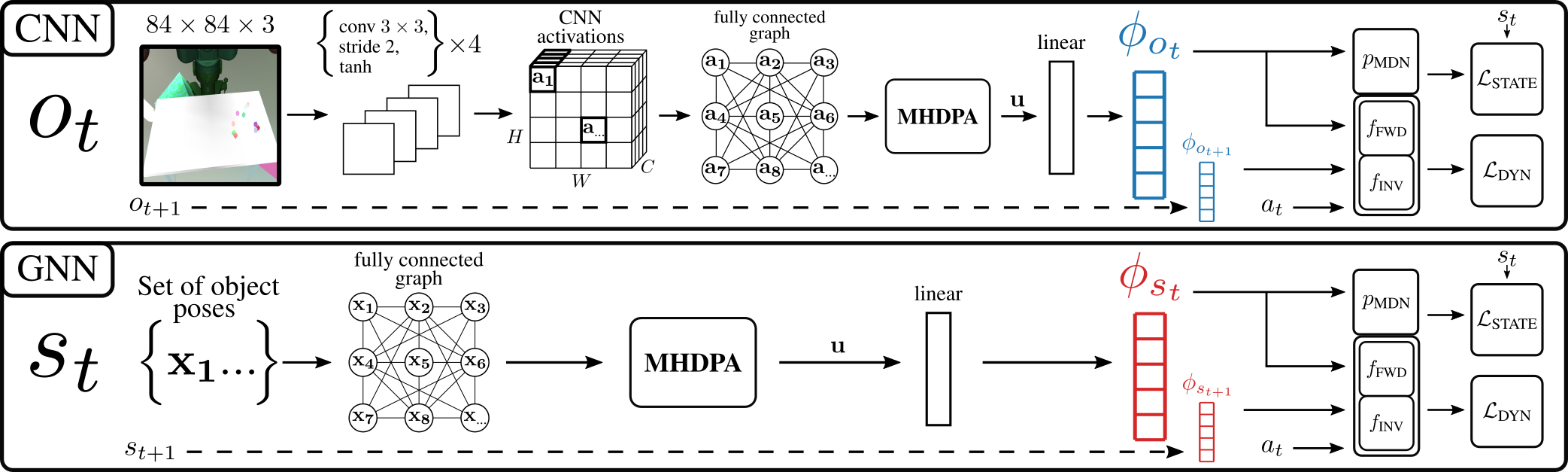}
	\caption[Architecture]{CNN and GNN architectures and losses for self-supervised representation learning}
	\label{fig:arch}
\vspace{-0.1in}
\end{figure}
If our state, $s$, consisted of a single object pose $\xx$,
we could simply learn our state encoder \(\Em_s\) via a multi-layer perceptron, $\phi = \textbf{MLP}(\xx)$.
To scale this approach to many objects, we could increase the input size of the MLP and set
unused values to zero when there are fewer than the maximum number of objects present.
Unfortunately, in practice, the weak inductive bias of this approach makes it poorly suited
for moderately large numbers of objects.
With a set of $n$ objects, there are $n!$ ways that these objects can be arranged: $(\xx_1, \xx_2, ..., \xx_n)$, $(\xx_n, \xx_{n-1}, ..., \xx_1)$, etc.
Each of these will appear unique to the MLP, requiring an exponential order of training examples for the network to learn in the worst case \cite{graph_nets}.
To scale to many objects, we instead use a graph neural network (GNN) that operates on set structures in an order-invariant way.

The specific graph network model we use
is a single encoder block from the transformer model~\cite{transformer}.
One of the transformer's primary mechanisms, the multi-head dot-product attention (MHDPA)
can be interpretted as implementing a special parallelized version of updates and aggregations over a fully connected graph \cite{graph_nets}.
(Others have demonstrated the use of MHDPA on reinforcement learning tasks like Star Craft II \cite{zambaldi18, alphastar}.)
We convert our state into a fully
connected graph $G$ associating a vertex, \(\vv_i\), with each object centroid, \(\xx_i\). We use the \textbf{MHDPA} operation to
compute an updated graph of the same shape, $G' = \textbf{MHDPA}(G)$.
To aggregate information from the nodes $\vv'_i$ of $G'$, we compute a
gated activation \cite{wavenet} on each
node $\vv'_i$ followed by an elementwise summation over the
set of all nodes. This yields a single vector
that summarizes the full graph information:
$\globals = \sum_{i=1}^{n} \text{tanh}(W_1 \vv'_i) \odot \sigma (W_2 \vv'_i)$, where \(\odot\) denotes element-wise multiplication. 
(A simpler aggregation without the gated activation: $\globals = \sum_{i=1}^{n} \vv'_i$ also works, just not as well.) 
We then linearly transform $\globals$ to achieve
the desired vector size of our latent space: $\phi = W_3 \globals$.

\vspace{-0.25em}
\paragraph{The Output Side - Learning State Representations:}
We face similar concerns in constructing an appropriate loss function for learning to predict a variable number of outputs.  For the single object or fixed-length state setting, we could use mean-squared error regression from the latent space $\phi$
to that object pose $\xx$: $\frac{1}{2}\small\lVert g(\phi) - \xx\rVert^2_2$ \cite{aac, openai18, zhang16}.} 
This approach, however, is prone to failure in the multi-object setting.
For our image-based encoder network, \(\Em_o\), it would be ambiguous which labels ($\xx_1, \xx_2, ...$)
belong to which of several visually indentical objects in the state;
in this case, the network has no way of knowing which object predictions
it should place in which of its arbitrarily assigned prediction slots ($\hat \xx_1, \hat \xx_2, ...$).

\begin{figure}[t]
	\centering\vspace{-0.25em}
	\includegraphics[width=\linewidth]{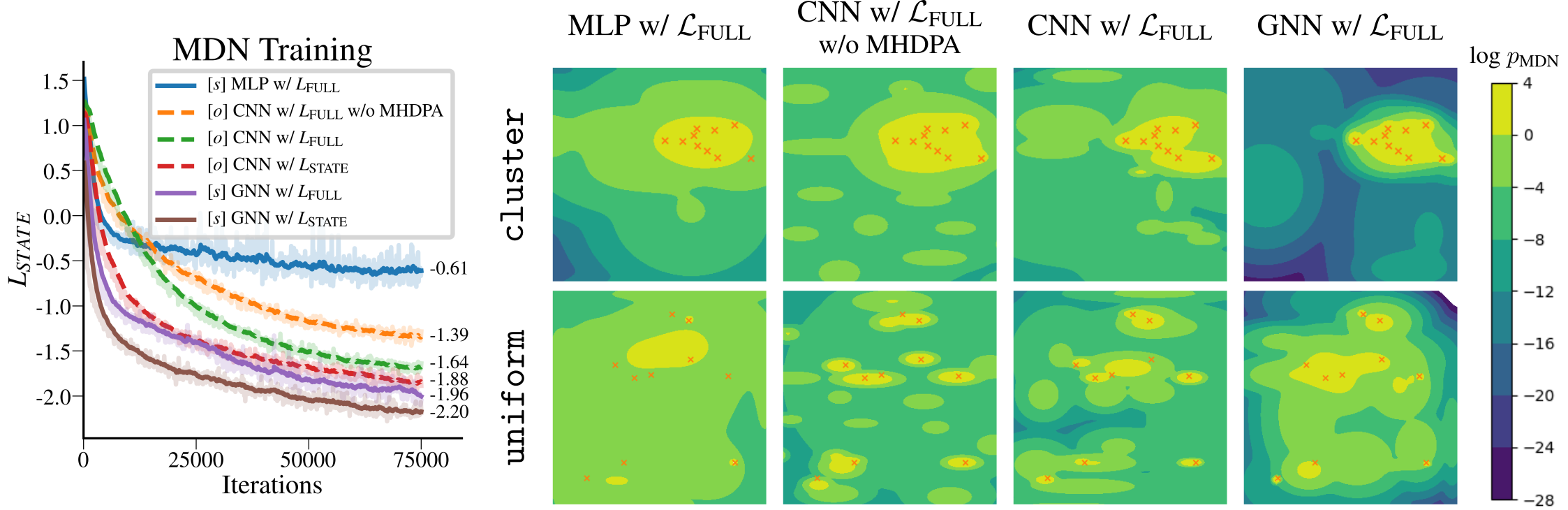}
	\caption[sim]{\textbf{Left:} MDN learning curves
	for various networks trained with either $\Lfull$ or 
	$\Lstate$. \textbf{Right:} Visualizations of the trained 
	MDNs; orange crosses represent ground truth locations.  All networks capture the state of clustered
	objects well, but the MLP fails to precisely
	localize the individual objects.
	Incorporating MHDPA into the CNN leads to fewer false positives and greater precision.
	And the GNN produces the tightest bounds and places the greatest amount of
	probability mass correctly on the objects and not other points.}
	\label{fig:mdn_training}
\vspace{-1em}
\end{figure}
To sidestep this and other similar issues, instead of producing individual predictions, we construct a mixture density network (MDN) \cite{mdn}. $p_\text{MDN}(\pp | \phi)$
encodes the probability of a 2D point $\pp$ on the table corresponding to one of the object centers $\xx_i$ conditioned on our latent representation \(\phi\):
\begin{equation}\label{eq:mdn}
	p_\text{MDN}(\pp | \phi) = \sum \limits_{k = 1}^{K} \alpha(\phi)_k ~\mathcal{N}\Big(\pp \Big| \mu(\phi)_k,~ \Sigma(\phi)_k\Big)
\end{equation}
The MDN is a Gaussian mixture model, with weights $\alpha_k$,
means $\mu_k$, and variances $\Sigma_k$, computed as
learned linear transformations of the latent space.
(See Fig.~\ref{fig:mdn_training} for visualizations of MDN predictions.)
We train $p_\text{MDN}$ using a maximum likelihood loss
which optimizes the network to assign high probability to
locations of ground truth object centers ($\xx_1, \xx_2, ..., \xx_n$):
\footnote{We note that our MDN loss is specific to tasks where object types
can be ignored.  One could instead use 
an object-type-sensitive loss or perhaps one 
which assigns probability density to space contained within objects.}
\begin{equation}\label{eq:mdn-loss}
	\Loss_\text{STATE} = - \frac{1}{n} \sum \limits_{i=1}^{n} \text{log}~p_\text{MDN} (\pp = \xx_i | \phi)
\end{equation}
While $\Lstate$ enables our network to learn a latent representation that can accurately capture object position information,
it does not account for dynamics information.
Some state encodings may be close according to $\Lstate$,
but difficult to traverse between, based on the actions
available to an agent.
To account for state traversability and actionability, we train our latent space using forward and inverse dynamics functions, as in prior work (e.g.~\cite{pokebot, icm}):
\begin{equation}\label{eq:dyn-loss}
	\Loss_\text{DYN} = \Loss_\text{FWD} + \Loss_\text{INV} = \frac{1}{2} \left\lVert f_{\text{FWD}}(\phi_{t}, a_t) - \phi_{t+1} \right\rVert^2_2 + \frac{1}{2}\left\lVert f_{\text{INV}}(\phi_t,\phi_{t+1}) - a_t\right\rVert^2_2
\end{equation}
Here $f_\text{FWD}$ is a small two-layer MLP trained to predict the forward dynamics in the latent space; while $f_\text{INV}$ is a small two-layer MLP trained to predict the inverse dynamics---the necessary action to move between two consecutive latent-space states. We use $\Lfull$ to train our encoders:
\begin{equation}\label{eq:full-loss}
	\Loss_\text{FULL} = \Loss_{\text{STATE}} + \Loss_{\text{DYN}}
\end{equation}

We hypothesized that the addition of $\Ldyn$ would help regularize the latent space and condition it for use 
in deciding actions, leading to better performance in the real world.
Our results seem to weakly support this hypothesis, with an $\Lstate$ model performing
slightly (but not significantly) better in simulation and an $\Lfull$ model generalizing slightly (but not significantly) better to the real world. Figure~\ref{fig:mdn_training} summarizes the training results for our encoder networks.

\paragraph{Architecture Summary:}
To summarize, we learn two separate encoders: 
$\mathbf{E}_\text{GNN} : s \rightarrow \phi_s $ which maps from state, %
and $\mathbf{E}_\text{CNN} : o \rightarrow \phi_o$ which maps from image observations. %
Each encoder learns separate forward and inverse dynamics functions as well 
as separate MDN parameters.\footnote{We first tried learning \(\phi_s\) and setting it as a regression target for our CNN encoder, but this failed to produce useful estimates. We also tried sharing the output dynamics and density weights, but this performed worse.} Fig.~\ref{fig:arch} shows the full architecture.
We incorporate an \textbf{MHDPA} operation into the CNN as in prior work~\cite{zambaldi18}.

As illustrated in Fig.~\ref{fig:arch},
we treat each of the ($1\times 1 \times C$ sized) slices of the last ($W\times H \times C$ sized)
CNN activation as elements of a set (of cardinality $N^{W \times H}$),
and compute the same \textbf{MHDPA} and aggregation operations
we use in the GNN.
Incorporating GNN mechanisms into CNN architectures this way
has been shown to avoid some shortcomings of vanilla CNN models in tasks
that require tracking object counts and reasoning about relative positions of objects \cite{rns}, and we also find it helps.

\subsection{RL Training}
\label{sec:rl}
During the RL phase, we use our grounded state representations
to train an image-based policy via an off-policy actor critic algorithm
(Soft Actor Critic (SAC) \cite{sac} in this case).
We use the doubly-grounded state-based representation $\phi_s$ to 
both compute rewards and train the critic efficiently.
We apply domain randomization and use the image-based representation $\phi_o$ 
to train the actor. %
Our full approach, which we call Multi-Object Asymmetric Actor Critic (MAAC), is described in Algorithm~\ref{algo:maac}.
\vspace{-1em}
\begin{algorithm}[h]
   	\footnotesize
   	\caption{MAAC: Learning manipulation tasks using a Multi-Object Asymmetric Actor Critic}
   	\label{algo:maac}
   	\begin{algorithmic}[1]
		\Require Encoders $\Ecnn(o_t)$, $\Egnn(s_t)$, replay buffer $R$, SAC \cite{sac} or other off-policy RL algorithm
		\State Collect dataset $\mathcal{D}$ in simulation using scripted policy, while applying domain randomization to images
    \State Train encoders $\Ecnn$ and $\Egnn$ on $\mathcal D$ by optimizing \eqref{eq:full-loss}
		\For{$\text{simulation episode~} e = 1,\ldots, M$}
		\State Sample and embed initial state $\phi_{s_0}=\Egnn(s_0)$, observation $\phi_{o_0} = \Ecnn(o_0)$, and goal $\phi_g = \Egnn(g)$

		\For{$t = 0,\ldots,T$}
			\State Execute action $a_t = \pi(\phio, \phig)$, and obtain new state $s_{t+1}$ and observation $o_{t+1}$
			\State Embed new state and observation $\phisn =\Egnnsn$, $\phion = \Ecnnon$
			\State Compute reward $r_t = r(\phis, \phig)$, and store transition $(\phis, \phio, \phig, a_t, r_t, \phisn, \phion)$ in $R$
		\EndFor
		\State Generate virtual goals ($g'_1, g'_2, ...$) and rewards ($r'_1, r'_2, ...$) for each step $t$ and store in $R$ (w/ HER \cite{her})
		\State Optimize actor ($\pi(\phio, \phig)$) using SAC with $o_t$ embeddings
		\State Optimize critic ($V(\phis, \phig), Q_1(\phis, \phig, a), Q_2(\phis, \phig, a)$) using SAC with $s_t$ embeddings
    \EndFor
   	\end{algorithmic}
\end{algorithm} 
\vspace{-1.5em}
\paragraph{Algorithm~\ref{algo:maac} Overview:}
First, before the RL phase, 
we run a scripted policy to collect a dataset $\mathcal{D}$ of transitions of 
states and observations ($s_t, o_t, a_t, s_{t+1}, o_{t+1}$) (line 1).
We use $\mathcal{D}$ to train our encoders via supervised learning (line 2).
Then, for each RL episode, we first embed the goal and initial observation and state (line 4).
At each step, we pass the observation embedding $\phio$ through the policy network
to get an action (line 6).
We also use the state embedding $\phis$ (along with $\phig$) to compute 
a reward (line 8), and later to feed to the value network.
We parameterize the policy network $\pi(\mathbf{\phio}, \phig)$ 
and the value networks $\left(V(\mathbf{\phis}, \phig), Q_1(\mathbf{\phis}, \phig, a), Q_2(\mathbf{\phis}, \phig, a)\right)$ by MLPs with two hidden layers with
the learned embeddings as input; we optimize them using standard SAC losses.

\paragraph{Specifying a Reward:}
To specify a reward for multi-object tasks, we face similar considerations
as when embedding a variably sized state (Section~\ref{sec:gsr}).
Our approach is to use the GNN-based latent space vector to specify
a distance between the current state and goal state.
To do this, we embed the current and goal states $\phis = \Egnns$ and $\phi_g = \Egnng$.
Then we specify a goal success condition $f_g(s_t)$ by computing
the cosine distance between these embeddings;
when the cosine distance falls below a threshold $\epsilon$, the goal
is considered met.  This can be described by the following equation:
$f_g(s_t) = \texttt{cos\_dist}(\phis, \phig) < \epsilon$.
For every step $t$ that the agent has not reached the goal,
we provide a $-1$ reward.  Once it reaches the goal we 
provide a $+1$ reward.  
\footnote{
We originally tried using a denser reward based on changes in cosine distance
to the goal, but found it did not work as well; we think likely because it penalized
incorrect movement too much, hindering exploration \citep{her}.}
We terminate episodes with a final $-1$ reward if the goal
condition has not been met after $T$ steps (where $T \approx 50$).

\paragraph{Tricks to Speed Up Training:}
We keep the encoder networks frozen in this phase and cache the latent embeddings, 
increasing the maximum batch size and size of replay buffer we can keep in memory and 
reducing the computation required to train the SAC heads. 
To incentivize object movement, we add an extra reward
of $+0.1$ when the agent moves any of the objects more than $\delta$ cm (where $\delta \approx 2.0$).
To improve exploration, we usually sample actions from our learned policy $a_t = \pi(\phio, \phig)$ (line 6), but
with probability $p$, we sample actions from our scripted policy.
We set $p$ to 1.0 early in training and linearly anneal it over time.
To help learn from our sparse reward, 
we use Hindsight Experience Replay (HER) \cite{her} to augment
the replay buffer (line 10).
Also, our actual implementation is a parallelized version of Algorithm~\ref{algo:maac}, where we
intermix SAC optimization cycles with experience collection. \footnote{We additionally tried directly learning a CNN policy \(\pi(O, \phi_g)\) without pre-training a latent representation from images, but found this to train much slower during RL as we could not use the cached embeddings.}
See the supplementary material for these implementation details.

\vspace{-0.2em}
\section{Experimental Results}
\label{sec:results}
\vspace{-0.8em}
In our experiments we study the following questions:
\begin{enumerate}
 \item How does the full method compare with alternative training formulations in simulation? (Sec.~\ref{sec:results_sim}) 
 \item What is the effect of domain randomization and using $\phi_s$ in policy training? (Sec.~\ref{sec:aac_domrand})
 \item How do the learned policies compare and generalize in real-world experiments? (Sec.~\ref{sec:results_real})
\end{enumerate}
We train our policies purely in simulation and evaluate them in simulation and the real world.
We use two methods for sampling states;
\texttt{uniform}: where all objects are uniform randomly
distributed across the table; and
\texttt{cluster}: where objects are sampled from
25cm x 25cm square areas on the table.
For initial state sampling $p(s_0)$,
we use \texttt{uniform} and \texttt{cluster} each 
50\% of the time.
For goal sampling, we always use the \texttt{cluster} approach.
We train with 10 cubes, but evaluate on various objects in the real world.
We compare our full method to several methods and ablations, including
an MLP and autoencoder, an image-based goal ($\pi(\phi_s, \phigo)$ vs. $\pi(\phi_s, \phig)$),
a CNN that does not incorporate the MHDPA mechanism, and models trained with only
$\Lstate$ or $\Ldyn$ rather than $\Lfull$. 

Unless otherwise specified, we always use
domain randomization and an asymmetric actor critic (AAC) approach---using $\phi_s$
for both computing rewards and value functions.
$\phi_s$ and $\phi_o$ are always vectors of size 128.
We train all models on $\Lfull$ (or some ablation of it), besides the autoencoder approach, which uses
a reconstruction loss.
\footnote{
We found that a naive autoencoder approach failed to train on this task when
applying domain randomization.  We had to apply a semi-novel technique,
based on prior canonicalization work \cite{rcan}, to get it to work at all.}
Recall that during RL training, we determine goal success as $f_g(s_t) = \texttt{cos\_dist}(\phis, \phig) < \epsilon$.
If $\epsilon$ is too high, the task is trivially easy, and
if it is too low, the agent will never reach the goal.
To ensure $\epsilon$ corresponds to approximately equal state similarities across approaches, 
we collect a set of transitions, compute a list of cosine distances for each model,
and apply a scaling based on the relative values that the models produce.

\begin{figure}[h!]
	\centering
	\includegraphics[width=0.99\linewidth]{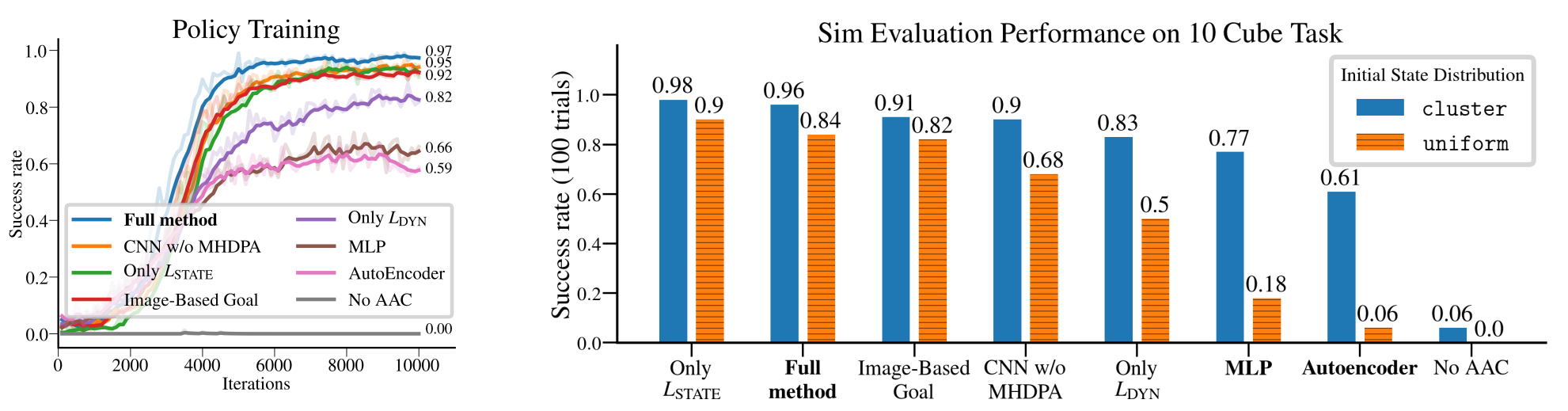}
	\caption[sim]{{  
	\textbf{Left:} Sim training. Our full method reaches a higher success rate in a fewer
	number of training iterations compared to an MLP approach,
	an autoencoder approach, and several ablations.
	\textbf{Right:} Evaluation results. We evaluate all trained models on a slight variation of the task
	that does not rely on thresholds in a learned embedding space.
	We find that the full method produces nearly the best performance,
	only being slightly beaten out by a model trained on $\Lstate$.
	We test models using both a clustered and uniform initial
	state distribution of objects, and find that other methods
	perform worse, especially with objects initially uniformly distributed.%
	}}
	\label{fig:results_sim}
\end{figure}

\vspace{-0.8em}
\subsection{Simulation Eval Performance}
\label{sec:results_sim}
\vspace{-0.8em}
To evaluate our training models in simulation, 
we run 100 trials of each method and count the success
rate of when they are able to push the cubes into the desired regions.
We cannot rely on the learned $\phi$ space to fairly compare
model performance, as some models could be more lenient of state 
similarities.  Instead we iterate through
the positions of all the objects and check if they fall within
the 25cm x 25cm goal area.
(See more detail in the supplementary material.)
The \texttt{uniform} distribution (which is sparse across the table) better represents true
multi-object handling, whereas a cluster of objects is easier and more
like a single-object setting.
The results shown in Fig.~\ref{fig:results_sim} illustrate that the MDN and full method best handle this more challenging setting.

\begin{figure}[t]
	\centering
	\includegraphics[width=\linewidth]{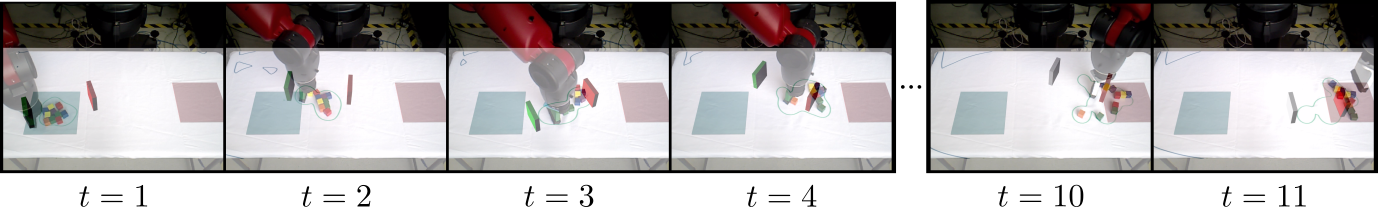}
	\caption[sequence]{Sequence of states and pushing actions chosen by a policy trained with our full approach leading to successful execution. The blue and red square overlays represent the initial state and goal areas sampled with the \texttt{cluster} state distribution. Contours represent the MDN output density estimates of object locations.}
	\label{fig:sequence}
\vspace{-0.15in}
\end{figure}

\vspace{-0.2em}
\subsection{Effects of Domain Randomization and Using $\phi_s$}
\label{sec:aac_domrand}
\vspace{-0.8em}
\begin{wrapfigure}{r}{.24\columnwidth}
\vspace{-14mm}
\centering
\includegraphics[width=0.24\columnwidth]{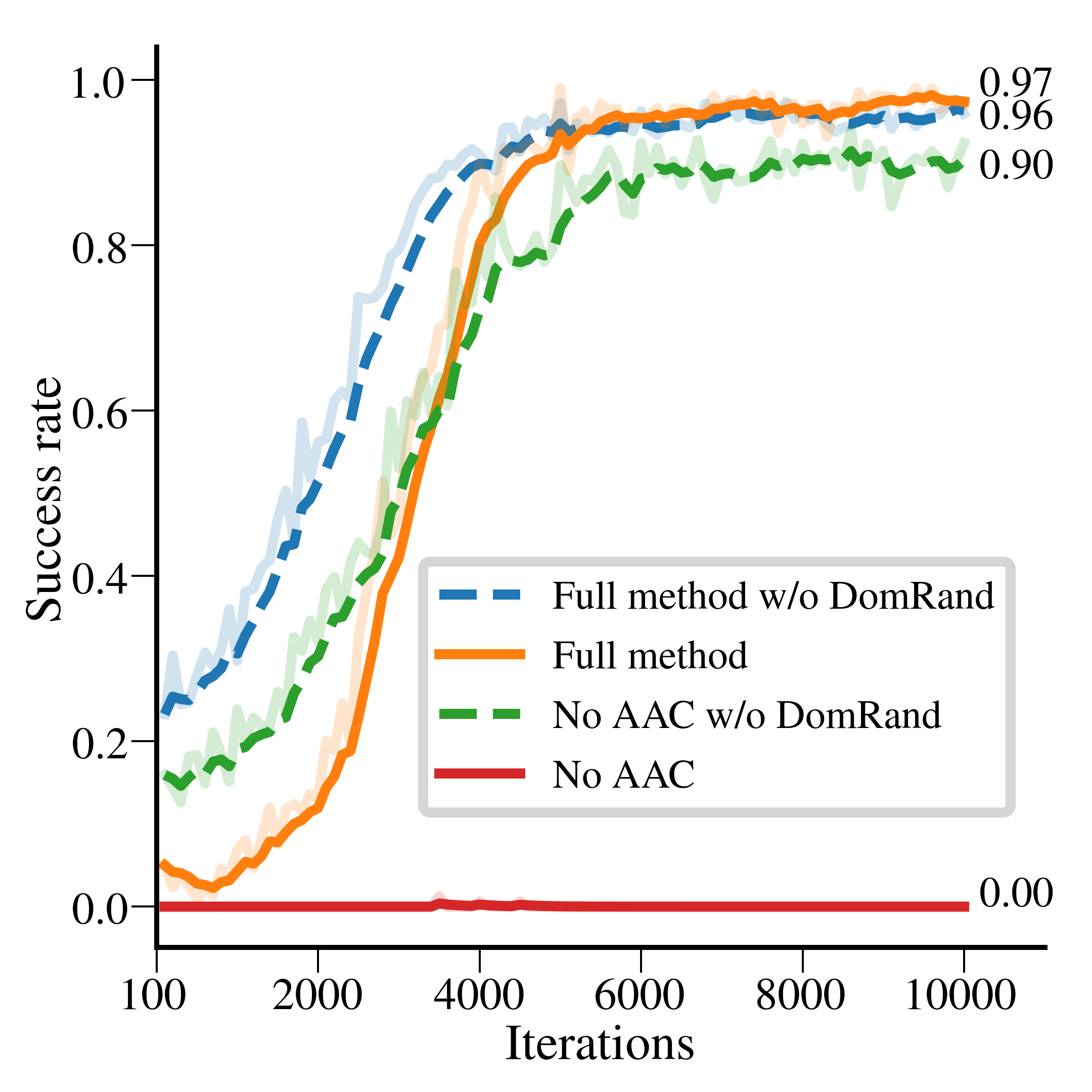}
	\caption{\small{Effects of domain randomization and asymmetric actor critic.}}
\label{fig:acc_domrand}
\vspace{-0.40in}
\end{wrapfigure}
We found that applying domain randomization (DomRand) made
the $\phi_o$ estimates noisy and highly variable, even when the underlying state
did not change.  This ultimately led to our use of $\phi_s$ in an asymmetric 
actor critic approach (AAC) to stabilize training signals.
To study these effects, we ran our full approach with different
combinations of AAC and DomRand. 
When applying DomRand, we find the use
of AAC makes the difference between the policy achieving high success
rates and completely failing to learn.  AAC seems to help primarily by stabilizing learning signals.

\vspace{-0.2em}
\subsection{Real World Evaluation Performance}
\label{sec:results_real}
\begin{figure}[h]
	\centering
	\includegraphics[width=\linewidth]{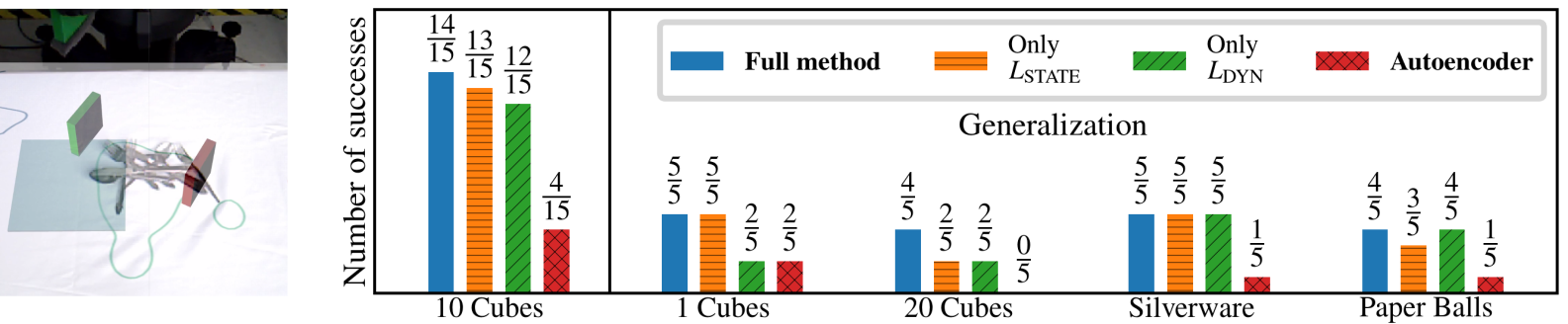}
	\caption[real]{\small{\textbf{Left:} Silverware task. 
	\textbf{Right:} Real world evaluation results. We evaluate on 15 different start states and goal locations for the 10 Cube training task, and on 5 for the other generalization tasks.
	We terminate after successes or 15 push actions.
	We hold start states and goals constant across different methods.
	The full method produces the greatest performance and generalizability, especially outperforming the autoencoder approach.}}
	\label{fig:results_real}
\vspace{-0.1in}
\end{figure}

\vspace{-0.8em}
We use a similar approach as we did in simulation for evaluating our policies in the real world, but we
always use \texttt{cluster} as our initial state distribution.
To study which loss function generalizes best,
we compare the full method, $\Ldyn$, $\Lstate$, and the autoencoder
as these all learn different image models.
Along with the 10 cube task we train on, we test on tasks with 1 cube; 20 cubes; 10 spoons, knives, and forks (Silverware);
and 5 crumpled paper balls. 
Fig.~\ref{fig:sequence} shows a successful sequence of states and pushing actions taken by the robot using the trained policy. Fig.~\ref{fig:results_real} shows quantitative results.
We see that all variants of our approach significantly outperform the autoencoder, and $\Lfull$ 
slightly outperforms the ablations in these trials. See the associated video for more experiment visualizations.

\vspace{-0.2em}
\section{Discussion}
\label{sec:discussion}
\vspace{-0.8em}
We present an approach for sim-to-real robot manipulation learning
that exploits variable and multi-object simulator state.
We do this by learning grounded state representations and using them to train an RL policy that outperforms alternative approaches in both simulated and real-world experiments.
\vspace{-0.7em}
\paragraph{Limitations and Future Work:}
We use a coarse goal specification of 25cm x 25cm areas. 
It would be interesting to try to achieve more precise goal states,
perhaps by gradually annealing the threshold $\epsilon$ during training. 
Our scripted policy is specific to the pushing task we consider, and the MDN 
loss ($\Lstate$) is specific to tasks where objects can be considered all of 
the same type.  Exploration-based policies (e.g., \cite{icm}) 
and set-based loss functions (e.g., \cite{zhang19}) could perhaps be used instead.
We tried \texttt{uniform} initial states in the real world, but found our policy
often got stuck.  We believe incorporating memory (e.g., with recurrent neural networks)
may lead to more adaptive and robust policies.

\clearpage
\acknowledgments{Matthew Wilson was supported in part by the National Science Foundation under Grant \#1657596, as well as the Undergraduate Research Opportunities Program at the University of Utah.}

\small{
\bibliography{sources}
}

\appendix

\section{Hardware and Software}

\paragraph{Hardware:}

\begin{wrapfigure}{r}{.3\columnwidth}
\vspace{-5mm}
\centering
\includegraphics[width=0.3\columnwidth]{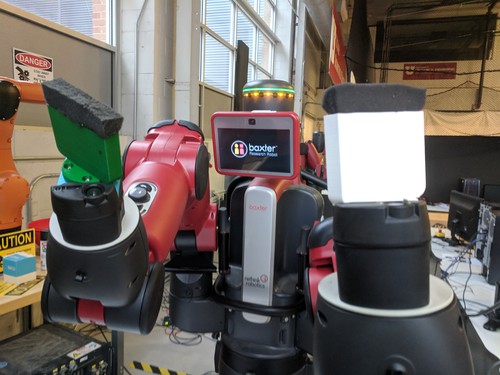}
	\caption{Close-up of compliant pushing end effectors.}
\label{fig:paintbrushes}
\vspace{-0.30in}
\end{wrapfigure}

All models for the full method were trained on a single desktop computer
(Intel i7-8700k CPU @ 3.70GHZ x 12) with an NVIDIA GTX 1080 Ti GPU.
Some of the comparison models were also trained on a shared NVIDIA DGX machine.
We use the Baxter robot from the late great RethinkRobotics \cite{baxter}, with
custom 3D-printed plastic end effectors and foam inserts for compliant pushing of objects.
We use the ASUS Xtion Pro RGB-D camera to capture images during
inference.  We only use the RGB channels, and we downscale
the $640 \times 480 \times 3$ image the camera produces
to size $84 \times 84 \times 3$ for use in our CNN.
We use the same render image size in simulation and
downscale the same way.

\paragraph{Software:}
We use the Mujoco simulator with its built-in renderer.
During training in simulation, for simplicity, we do not articulate the robot arm
to push the objects.  Instead, we articulate a disembodied paddle using simple
PD control (seen in Fig.~\ref{fig:task}).  
We also place invisible walls around the table to prevent
objects from falling off. 
To disincentivize policies from learning strategies 
that would push objects off the table in the real world,
we give -1 reward each time one of the objects contacts the walls.
In the real world, we use the Move-It! manipulation software, 
with an RRT Connect planner to plan pushing actions for the 7-dof Baxter arms.
We choose to either use the right or left arm based on where the pushing action is on the table.
During our evaluation, we run an autonomous loop where we: (1) collect an image
from the ASUS camera, (2) feed that image to the policy, (3) sample and execute an action, and (4) repeat until
15 pushes have been executed or the goal has been reached.
During training, we always sample actions according to the Gaussian parameterized by the SAC policy network.
At inference, we continue to sample, but we reduce the standard deviation by 50\%.  We find this strikes
a good balance between having too much noise, where the policy often executes suboptimal pushes,
and having too little noise, where the policy gets stuck in certain states when slightly confused about the location of the objects.

We use TensorFlow as our deep learning framework.
On top of that, we use TensorFlow Probability \footurl{https://www.tensorflow.org/probability}
to implement our Mixture Density Network.
And we use the Sonnet \footurl{https://github.com/deepmind/sonnet} and graph\_nets \footurl{https://github.com/deepmind/graph_nets}
frameworks developed by DeepMind to implement our CNNs
and GNNs, including by using their implementation of multi-head dot product
attention (MHDPA).

\section{Implementation Details}

\paragraph{Dataset for Training the Encoders:}
We collect 1 million transitions in our initial dataset for training the model.
This takes about 24 hours of wall time on the desktop machine.
We did not experiment with using less data, but we suspect 1 million is
a bit overkill; fewer examples may have worked just as well.
We also apply data augmentation by adding random image contrast (\texttt{tf.image.random\_contrast}) and noise (\texttt{tf.random.normal}) to the image as well as the embeddings before applying the MDN and dynamics heads.

\paragraph{Autoencoder Details:}

Our Autoencoder encodes its input into a latent space $\phi$
and then outputs a set of logits of the same size as the observation
image.  To better control for differences in the model,
we use the same asymmetric actor critic apporach we do 
with the full method.  We use both a CNN and GNN that map to
latent spaces $\phi_o$ and $\phi_s$, repsectively.  
We then independently pass both these latent spaces
through a series of upconvolutions 
to output a 3D array ($W \times H \times C$) of logits that parameterize
independent Bernoulli distributions
which are trained to maximize the log probability of the true pixel values.
The difference between this and a full Variational Autoencoder is
that we do not enforce a prior distribution onto the latent 
space and we do not sample from the latent space.  We tried enforcing this prior and sampling, 
but found that doing this did not allow the model to capture the object positions, at least during initial development.
We use the same architecture for the GNN and CNN as we do in our full method.
We find that the GNN-based representation provides a similar
performance boost as what we see in the full method.

\begin{wrapfigure}{r}{.4\columnwidth}
\vspace{-10mm}
\centering
\includegraphics[width=0.4\columnwidth]{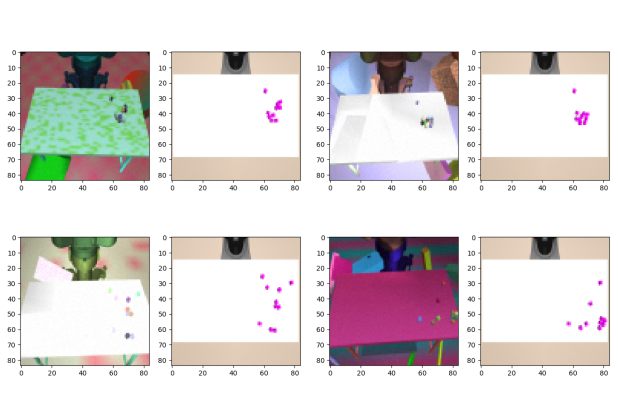}
	\caption{Domain randomized (1st \& 3rd columns) and canonical (2nd \& 4th) observations.
	Domain randomized images are used for the inputs of the
	Autoencoder, whereas a standard canonicalized version of the
	image is always used for the output prediction.  This canonicalized
	version is always from an overhead view with no background
	distractors and only majenta objects.}
\label{fig:canonical}
\vspace{-17mm}
\end{wrapfigure}

Initially, we found that applying domain randomization led this Autoencoder approach 
to fail. All of the capacity of the Autoencoder was wasted
modeling the randomized textures and varying camera characteristics
and it was not able to localize the objects.
To solve this issue, we use the canonicalization approach
from \citet{rcan}.  The idea is to collect
another image observation at each time step that is in a ``canonical" format
which is not randomized and thus the only source of difference is the
object positions (see Fig.~\ref{fig:canonical}).

\paragraph{MLP Details:}
Our MLP consumes the set of $n$ object coordinates as an array of size $n \times 2$ (which
gets flattened) and produces a latent representation $\phi$.
We design it to have roughly the same number of parameters
as the GNN, as shown in Table~\ref{tab:model_params}.  
It consists of 3 hidden layers of size 256.

\section{Hyperparameters}
We use the same hyperparameters, except for $\epsilon$,
to train all of the models
during the supervised and reinforcement learning phases.
We found the GNN and MLP were robust to different learning
rates, so we use a value which works well for the CNNs,
including the Autoencoder network.  We find using
fairly large batch sizes was helpful to learn while
applying noise via domain randomization and data augmentation.
We use tanh (instead of ReLU) activations because prior sim-to-real work 
suggests that since tanh is a smoother activation, it leads
to better domain transfer performance \cite{anymal}.

We list the supervised learning hyperparameters in Table~\ref{tab:slhps},
the reinforcement learning hyperparameters in Table~\ref{tab:rl_hps},
and the model parameter counts in Table~\ref{tab:model_params}.
The asterisks (*) represent hyperparameters from the supervised phase
that stay the same in the RL phase.

\begin{table}[h]
	\centering
	\caption{\normalsize Supervised representation learning hyperparameters to train all models}
	\label{tab:slhps}
	\vspace{2mm}
	\begin{tabular}{cc}
		\hline 
		Hyperparameter & Value \\ 
		\hline
		optimizer* & Adam \\ 
		learning rate & $3 \times 10^{-4}$ \\ 
		batch size & $512$ \\ 
		number of iterations & $75000$ \\ 
		activation* & tanh \\ 
		conv filters* & $64$ \\ 
		$\phi$ size* & $128$ \\ 
		MDN $K$* & $25$ \\ 
		transformer vector size* & $128$ \\ 
		transformer num heads* & $8$ \\ 
		\hline 
	\end{tabular} 
\end{table} 

\vspace{-0.2in}

\begin{table}[h]
	\centering
	\caption{\normalsize RL hyperparameters to train all models}
	\label{tab:rl_hps}
	\vspace{2mm}
	\begin{tabular}{cc}
		\hline 
		Hyperparameter & Value \\ 
		\hline 
		learning rate & $1 \times 10^{-3}$ \\ 
		batch size & $1024$ \\ 
		replay buffer size & $1 \times 10^6$ \\ 
		$N$ (max episode steps before termination) & $50$ \\ 
		number of parallel environments & $8$ \\ 
		number of iterations & $10000$ \\ 
		$\phi$ noise & $0.1$ \\ 
		HER \cite{her} k & $8$ \\ 
		SAC \cite{sac} entropy bonus ($\alpha$) & $0.1$ \\ 
		polyak averaging ($\rho$) & $0.995$ \\ 
		\hline 
	\end{tabular} 
\end{table} 

\vspace{-0.2in}

\begin{table}[h]
	\centering
	\caption{\normalsize Number of parameters in different network models (during supervised learning)}
	\label{tab:model_params}
	\begin{tabular}{c|c}
		\hline 
		Model & Number of parameters \\ 
		\hline 
		GNN for Full & 232197 \\ 
		CNN for Full & 353157 \\ 
		CNN Autoencoder and Decoder & 937975 \\ 
		MLP (3 hidden layers of size 256) & 253445 \\ 
		\hline 
	\end{tabular} 
\end{table} 

\begin{table}[h]
	\centering
	\caption{\normalsize $\epsilon$ values}
	\label{tab:eps_vals}
	\begin{tabular}{c|c}
		\hline 
		Model & $\epsilon$ Value \\ 
		\hline 
		Full & 0.005\\ 
		Autoencoder & 0.2\\ 
		Only $\Lstate$ &  0.04\\ 
		Only $\Ldyn$ & 0.004\\ 
		MLP & 0.005\\ 
		\hline 
	\end{tabular} 
\end{table} 

\paragraph{Tuning $\epsilon$ Across Methods:}
To ensure that all methods face a similarly difficult training
task, we have to make sure the $\epsilon$ we use correspond to
similar state similarities.
We do this empirically by collecting a set of $4096$ transition pairs
($s_t, s_{t+1}$) using our scripted policy.  We filter out all
transitions where the objects have not moved at all.
Then we compute the cosine distances between all
remaining pairs for each of the models.  
Finally, we compute relative scalings between the values produced by each model.
Through the manual tuning of $\epsilon$ for various models,
we found these relative scalings correlated well with similar
task difficulties and training performances.
The resultant $\epsilon$ values are shown in Table~\ref{tab:eps_vals}.

\section{Extra Background}

\subsection{Soft (Asymmetric) Actor Critic:}
To train a goal-conditioned policy, we use a slightly modified version of the Soft Actor Critic (SAC) algorithm \citep{sac}.
SAC is a Deep Deterministic Policy Gradient (DDPG) style algorithm,
which optimizes a stochastic policy in an off-policy way.
It is based on the maximum entropy reinforcement learning framework \cite{ziebart08}
and has been efficient and robust for learning real-world robotics tasks.
As opposed to vanilla SAC, we feed the value networks $(V, Q_1, Q_2)$ a representation encoded from the underlying state, \(s\), while we feed the policy network $\pi$ a representation encoding the observation (RGB image), \(o\), in an asymmetric actor critic style approach \cite{aac}.
We have found that it leads to
stable training and helped policies from plateuaing too early in training,
compared to other model-free RL algorithms we tried.

\subsection{Hindsight Experience Replay:}
We also use Hindsight Experience Replay (HER) \cite{her}.
HER is a method for augmenting replay buffer data by relabeling goals
to aid training in sparse reward settings.
After sampling a goal $g$ and collecting an episode of experience, HER
stores the original transitions $(s_t, a_t, r_t, s_{t+1}, g)$ in the replay buffer, as usual.
Additionally, for each step $t$, it samples a set of virtual goals $(g'_1, g'_2, ...)$ that
come from future states of the episode (which it has guaranteed to have reached), and it uses these to create several
relabeled transitions $(s_t, a_t, r'_t, s_{t+1}, g_i')$ that it also stores in the replay buffer.
The denser rewards of these relabeled transitions is used
to bootstrap learning and make the policy more capable of reaching original goals.

\end{document}